%% file: main.tex
\documentclass{SCIS2025}

\usepackage{makecell}
\usepackage{multirow}
\usepackage[utf8]{inputenc} % allow utf-8 input
\usepackage{threeparttable}
\usepackage{hyperref}       % hyperlinks
\usepackage{url}            % simple URL typesetting
\usepackage{booktabs}       % professional-quality tables
\usepackage{amsfonts}       % blackboard math \usepackage{generic}
\usepackage{cite}
\usepackage{amsmath,amssymb,amsfonts}
\usepackage{algorithmic}
\usepackage{graphicx}
\usepackage{algorithm,algorithmic}
\usepackage{hyperref}
\hypersetup{hidelinks=true}
\usepackage{textcomp}

\usepackage[normalem]{ulem}

\usepackage{caption}
\usepackage{pgfplots}
\usepackage{pgf-pie}
\usepackage[most]{tcolorbox} % 推荐加载 most，功能全
\usepackage{tikz}            % 若用 TikZ 图标
\usepackage{xcolor}          % 定义自定义色
\usepackage{subcaption}
\definecolor{midnightgreen}{rgb}{0.0, 0.29, 0.33}
\definecolor{deepgreen}{HTML}{0aa344}
\definecolor{deeppurple}{HTML}{7030a0}
\definecolor{deepblue}{HTML}{171d91}
\definecolor{brown}{HTML}{843c0c}
\definecolor{shadered}{HTML}{ffe5e5}
\definecolor{shadegreen}{HTML}{e5f7ed}
\definecolor{msftBlack}{RGB}{0,0,0}
\definecolor{lightred}{RGB}{255,163,163}
\definecolor{deepred}{RGB}{146,0,0}
\definecolor{cb1}{RGB}{251,180,174} % 柔粉
\definecolor{cb2}{RGB}{179,205,227} % 柔蓝
\definecolor{cb3}{RGB}{204,235,197} % 柔绿
\definecolor{cb4}{RGB}{222,203,228} % 柔紫
\definecolor{cb5}{RGB}{254,217,166} % 柔橙
\definecolor{cb6}{RGB}{255,255,204} % 柔黄
\definecolor{cb7}{RGB}{229,216,189} % 柔棕
\definecolor{cb8}{RGB}{253,218,236} % 柔粉紫
\definecolor{lightgraybox}{RGB}{245, 245, 245}

\usepackage[tikz]{bclogo}
\newcommand{\smallbcetoile}{\scalebox{0.7}{\bcetoile}}
\newcommand{\finding}[2][]{%
  \begin{bclogo}[
    couleur=msftBlack!05, epBord=1, arrondi=0.2, 
    logo=\smallbcetoile, marge=4, ombre=true, blur, 
    couleurBord=msftBlack!10, tailleOndu=1, 
    sousTitre={\em #1}
  ]{}
    #2
  \end{bclogo}
}

\begin{document}

\ArticleType{RESEARCH PAPER}
%\SpecialTopic{}
%\luntan
\Year{2025}
\Month{}
\Vol{68}
\No{}
\DOI{}
\ArtNo{000000}
\ReceiveDate{}
\ReviseDate{}
\AcceptDate{}
\OnlineDate{}
\AuthorMark{}
\AuthorCitation{}
%%%%%%%%%%%%%%%%%%%%%%%%%%%%%%%%%%%%%%%%%%%%%%%%%%%%%%%

%%% title: ±êÌâ
%%%   \title{title}{title for citation}
\title{Knowledge-Driven Agentic Scientific Corpus Distillation Framework for Biomedical Large Language Models Training}{Knowledge-Driven Agentic Scientific Corpus Distillation Framework for Biomedical Large Language Models Training}
\author[]{Meng Xiao$^{1,3}$, Xunxin Cai$^{1,2}$, Qingqing Long$^{1}$, Chengrui Wang$^{1}$, \\Yuanchun Zhou$^{1,2}$\& Hengshu Zhu$^{1}$}{{hszhu@cnic.cn}}
\address[1]{Computer Network Information Center, Chinese Academy of Sciences, Beijing, China}
\address[2]{University of Chinese Academy of Sciences, Beijing, China}
\address[3]{Duke-NUS Medical School, National University of Singapore, Singapore}
\input{0.abstract}


\maketitle

\section{Introduction}
\input{1.introduction}

\section{Related Work}
\input{2.related}

\section{Methods}
\input{3.method}

% \section{Results}
\section{Results}
\input{4.experiment}

\section{Conclusion}
\input{5.conclusion}

\Acknowledgements{
This work is partially supported by the National Natural Science Foundation of China (No. 62506351, No.92470204), the Beijing Municipal Natural Science Foundation (No.4254089). Hengshu Zhu is the corresponding author. 
}

% \section{Limitations}
% \clearpage
% \balance
\bibliographystyle{unsrt}
\bibliography{ref}

\end{document}

%% file: 0.abstract.tex
\abstract{Corpus distillation for biomedical large language models (LLMs) seeks to address the pressing challenge of insufficient quantity and quality in open-source annotated scientific corpora, which remains a bottleneck for effective LLM training in biomedical research.
This paper proposes a knowledge-driven, agentic framework for scientific corpus distillation, tailored explicitly for LLM training in the biomedical domain, addressing the challenge posed by the complex hierarchy of biomedical knowledge. 
Central to our approach is a collaborative multi-agent architecture, where specialized agents—each guided by the Medical Subject Headings (MeSH) hierarchy—work in concert to autonomously extract, synthesize, and self-evaluate high-quality textual data from vast scientific literature.
This agentic framework collectively generates and refines domain-specific question-answer pairs, ensuring comprehensive coverage and consistency with biomedical ontologies while minimizing manual involvement.
Extensive experimental results show that language models trained on our multi-agent distilled datasets achieve notable improvements in biomedical question-answering tasks, outperforming both strong life sciences LLM baselines and advanced proprietary models.
Notably, our AI-Ready dataset enables Llama3-70B to surpass GPT-4 with MedPrompt and Med-PaLM-2, despite their larger scale. Detailed ablation studies and case analyses further validate the effectiveness and synergy of each agent within the framework, highlighting the potential of multi-agent collaboration in biomedical LLM training.\footnote{The code supporting the findings of this study is available at \href{https://www.dropbox.com/scl/fo/c4osaktg0jaltf9q3ma6j/AAbK99-rjnzttUk9Hkf1G8E?rlkey=oon1lkdr8mon953drhj1v6iou&st=yqld7z36&dl=0}{DropBox}.}}

%% file: 1.introduction.tex
% \section{Introduction}
% \section{Introduction}
% Before the introduction of large language models (LLMs), biomedical NLP relied on rule-based systems and handcrafted feature engineering\citep{intro-1-1}, operating within limited feature spaces that hindered the utilization of semantic information and context\citep{intro-1-2}, significantly constraining the efficiency and accuracy of data-driven decision-making\citep{intro-1-3}.

\begin{figure}[!t]
    \centering
    % \vspace{-0.2cm}
\includegraphics[width=0.65\textwidth]{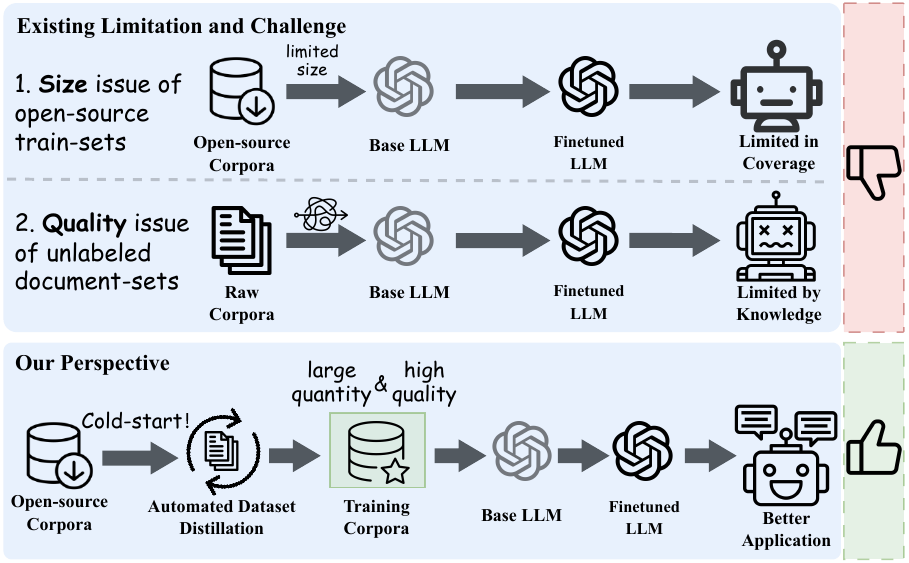} 
    \caption{Analyzing the limitations and challenges of the existing pipeline. The motivation of this study is to utilize the high-quality but limited annotated corpus to generate large-scale training corpora from raw scientific documents.}
    % \vspace{-0.8cm}
    \label{fig:Fig_intro}
\end{figure}

The advent of large language models has propelled bioinformatics into a new era~\cite{xi2025rise}, enabling the development of automated solutions across a spectrum of biomedical domains~\cite{luo2025intention, intro-1-1, intro-2-1,10878496}, and has demonstrated notable success for real-world biomedical question answering (QA) tasks~\cite{intro-2-2,wang2024biorag,xie2017topic}.
However, the intricate and specialized nature of biomedical tasks means that general-purpose LLMs often fall short unless meticulously adapted and fine-tuned for the domain~\cite{intro-3-3,icde1,qin2025scihorizon}.
Progress in this area is further constrained by the scarcity of sufficiently large and high-quality biomedical corpora~\cite{intro-1-2, intro-1-3, ai2025distillingclosedsourcellmsknowledge}.
While existing open-source biomedical datasets are typically of high quality, their limited scale and narrow topical coverage restrict their utility for comprehensive LLM training.
Conversely, directly leveraging the expansive body of domain-specific scientific literature offers the potential to cover a broader range of biomedical topics.
Nevertheless, most of these resources are unannotated, and their inherent lack of structure, coupled with the absence of QA-format organization, greatly hinders their effective use for training question-answering models.
Although the vast body of biomedical literature represents a valuable and authoritative resource, its complex terminology and dense conceptual structures pose significant barriers to automated processing and dataset construction~\cite{intro-3-2,intro-3-1}.
Figure~\ref{fig:Fig_intro} visualizes the resulting data bottleneck.
These challenges raise a crucial question: \textbf{How can we automatically distill large-scale, high-quality QA pairs from scientific literature to empower biomedical LLM?}

Prior research can be categorized into three mainstream categories. 
First, rule-based methods\cite{labrak2024biomistral,data_process-1,yangdistribution} rely on human-crafted standards for data cleaning and curation, which, while effective in reducing noise, are resource-intensive and difficult to scale. 
Second, knowledge graph-based approaches\cite{wu2024pmc,intro-prior-kg} structure biomedical information from texts into comprehensive graphs, but their dependence on manually curated sources limits both efficiency and scalability. 
Third, synthetic data generation methods\cite{medsyn-intro-prior-related-syn-2,med-text-mining,cai2023resolving,icde2,xiao2025interdisciplinary} use LLMs to automate the creation of QA pairs and process large corpora. Yet, they often lack mechanisms for interdisciplinary collaboration\cite{10052697,intro-isolate-source-bad}, resulting in insufficient diversity and robustness in the distilled data. 
Recent efforts~\cite{cai2025knowledge} have introduced a knowledge hierarchy-guided~\cite{xiao2021expert} approach that leverages a single LLM agent to generate and evaluate biomedical QA data with improved alignment to domain ontologies. 
Its reliance on a single-agent, rule-based architecture presents inherent limitations in terms of diversity, cross-domain expertise, and collaborative reasoning~\cite{luo2025large}.

In response to these limitations, we introduce the \textbf{M}ulti-agent enhanced \textbf{K}nowledge hier\textbf{A}rchy gu\textbf{I}ded biomedical dataset disti\textbf{L}lat\textbf{I}o\textbf{N} (\textbf{m-KAILIN}) framework—a novel, fully automated, agentic framework designed to extract high-quality training corpora for biomedical LLMs.
The m-KAILIN framework leverages a collaborative architecture of specialized LLM agents, each guided by a structured biomedical knowledge hierarchy, to evaluate and generate ‘Question-Answer-Context’ triples with strong domain alignment. 
The workflow begins with fine-tuned LLM agents generating candidate questions from limited annotated datasets. 
These questions are then matched to relevant contexts retrieved from over 23 million biomedical research articles. 
The question-context pair that best aligns with the knowledge hierarchy is chosen through a multi-agent evaluation and selection process. 
This automated pipeline facilitates the generation of high-quality, preference-based datasets that can further train LLMs to produce questions and answers aligned with biomedical ontologies, culminating in a robust, AI-ready corpus.

In summary, our contributions are listed as follows:

\begin{itemize}

    \item \textbf{Automated Biomedical Corpus Distillation Workflow}: We present a multi-agent workflow that enables the fully automatic extraction and distillation of biomedical corpora from large-scale research literature, vastly reducing the need for manual annotation while improving efficiency and dataset coverage.

   \item \textbf{Framework and Methodology}: Our framework introduces a knowledge hierarchy-driven evaluation mechanism, leveraging MeSH to guide and assess the multi-agent distillation. 
   This ensures the extracted data is both contextually relevant and domain-consistent, eliminating the need for human curation.

   \item \textbf{Comprehensive Empirical Validation}: We conduct extensive experiments to validate the effectiveness of m-KAILIN and the quality of the generated datasets. 
   Detailed ablation studies and case analyses reveal the contributions of each framework component, and our investigation of data scaling laws provides actionable insights into corpus distillation strategies for biomedical LLMs.
   
    % \item \textbf{Robust Knowledge Coverage}: Our experimental analysis demonstrates that m-KAILIN consistently performs well across a diverse range of biomedical topics. 
    % This robust knowledge coverage ensures the extraction process captures comprehensive, nuanced domain information, further enhancing the overall quality and adaptability of the generated corpus.
\end{itemize}

% These limitations highlight the need for efficient biomedical textual data mining methods that integrate knowledge hierarchy while preserving the integrity of the knowledge structure. 

%% file: 2.related.tex
\subsection{Dataset Distillation.}
Dataset distillation\cite{DD,fang2025knowledge} is an information extraction technique that uses generative models to distill core information from large-scale raw datasets, resulting in high-quality datasets.
Existing dataset distillation approaches are primarily categorized based on their optimization objectives into three types: performance matching, parameter matching, and distribution matching.
Performance matching methods \cite{DD_performance_1,DD_performance_2,DD_performance_3,DD_performance_4} aim to optimize dataset distillation by minimizing the loss of a model trained on the distilled dataset when evaluated on the raw dataset.
Parameter matching methods\cite{DD_parameter_1,DD_parameter_2,DD_parameter_3,DD_parameter_4} focus on the consistency of trainable model parameters when trained on the distilled dataset compared to the raw dataset.
Unlike these two training-based objectives, distribution matching methods\cite{DD_distribution_1} analyze the distilled dataset itself, using the distributional consistency between the distilled and raw datasets as the target for optimization.

% \textbf{Biomedical Synthesis Data:}
\subsection{Synthesis Text Data in Biomedical field.}
Existing methods for generating and utilizing synthetic text data in biomedical text mining face several challenges. 
PubMedQA \cite{jin2019pubmedqa} relies on converting titles into questions, often introducing noise due to simplistic or irrelevant content. 
Attempts to improve PubMedQA through LLM-based rewriting strategies, like those using GPT-3.5-turbo or GPT-4, fail to offer significant diversity in generated content\cite{improve-pubmedqa}. 
MedSyn\cite{medsyn-intro-prior-related-syn-2}, while leveraging a Medical Knowledge Graph and LLMs like GPT-4, lacks a comprehensive integration of hierarchical knowledge. 
Some other methods that rely on experts to use LLMs with their own knowledge~\cite{twelve-tips,med-text-mining} place too much dependence on expert effort, lacking automation and efficiency.
Besides, KAILIN~\cite{cai2025knowledge} first introduced the MeSH knowledge hierarchy as a rule-based evaluator, enhancing the quality of the generated questions. 
In this study, we advance beyond KAILIN by proposing an agentic, collaborative framework, where multiple specialized agents—each guided by biomedical ontologies—autonomously generate, evaluate, and refine the generated dataset.

\subsection{LLMs in Biomedical Domain.}
% \textbf{LLMs in Biomedical Domain:}
Recent advancements~\cite{10614318,10767279} in biomedical language models have focused on improving domain adaptation to better handle dense terminology and complex medical concepts. 
Approaches such as retrieval-augmented generation have enhanced models' ability to retrieve relevant information for domain-specific tasks\cite{intro-4-1,intro-4-2,intro-4-3}. PMC-LLaMA\cite{wu2024pmc} has been fine-tuned using millions of biomedical papers and medical textbooks, integrating medical knowledge graphs like UMLS for domain knowledge. Similarly, BioMistral\cite{labrak2024biomistral} has applied extensive data processing and adaptive pretraining on PubMed Central, while HEAL\cite{HEAL} has combined public and proprietary datasets for continuous pretraining on a general-purpose model, further enhancing its biomedical capabilities.

%% file: 3.method.tex
In this section, we detail the m-KAILIN framework, which leverages an agentic framework design to distill biomedical corpora from large-scale literature systematically. 

\subsection{Question Generation Agent}
The Question Generation Agent is responsible for transforming biomedical documents into high-quality, domain-specific questions. This agent operates in two stages: (1) fine-tuning large language models for biomedical question generation, and (2) utilizing the trained models to generate candidate questions from unseen documents.

\subsubsection{Fine-Tuning on Open-Source QA Dataset}\label{finetune-os}
To enable effective biomedical question generation, we fine-tune a large language model on the BioASQ QA dataset~\cite{bioasq-3}, and resulting in a specialized generator, $\theta$. Let $\mathcal{T} = \{(d_i, q_i)\}_{i=1}^N$ denote the open-source QA training set, where $d_i$ is a document and $q_i$ its reference question. The objective function is:
\begin{equation}
    \mathcal{L}_{\text{QA}}(\theta) = - \frac{1}{N} \sum_{i=1}^N \log P_{\theta}(q_i | d_i),
\end{equation}
where $P_{\theta}(q_i | d_i)$ is the model's predicted probability of the target question $q_i$. This cross-entropy loss ensures $\theta$ learn to generate questions aligned with biomedical knowledge.

\subsubsection{Question Generation from General Biomedical Documents}
During inference, each trained question generator receives a biomedical document $d_i$ sampled from PubMed and generates a corresponding candidate question $q_i$. Formally,
\begin{equation}
    q_i = \arg\max_{q} P_{\theta}(q_i|d_i)
\end{equation}
where $\theta$ denotes the fine-tuned question generator. This process enables the framework to automatically produce diverse and domain-relevant questions from large-scale biomedical literature, laying the groundwork for subsequent context retrieval and evaluation stages.

\subsection{Context Retrieval Agent}
The Context Retrieval Agent retrieves relevant supporting documents from the large-scale PubMed\footnote{PubMed: \href{https://pubmed.ncbi.nlm.nih.gov/}{Link}} corpus for each generated question. This agent employs Dense Passage Retrieval (DPR)~\cite{DPR}, in line with Retrieval-Augmented Generation (RAG)~\cite{RAG} frameworks. For each question $q_i$, we encode both the question and every candidate document $c$ as dense vectors, $\mathbf{e}(q_i)$ and $\mathbf{e}(c)$, using a shared BiomedBERT\textsubscript{base} encoder\footnote{BiomedBERT: \href{https://huggingface.co/microsoft/BiomedNLP-BiomedBERT-base-uncased-abstract-fulltext}{Link}}. The relevance score is computed as:
\begin{equation}
    p(c|q_i) \propto \exp\left(\mathbf{e}(c)^\top \mathbf{e}(q_i)\right),
\end{equation}
Documents with top-$k$ relevance scores are selected as contexts $\textbf{c}_i=\{c_1,c_2,...,c_k\}$ for subsequent evaluation. This decoupled retrieval enables independent optimization and clearer agent collaboration in the overall framework. 

\subsection{Question Evaluation Agent}
The Question Evaluation Agent is designed to assess which of two candidate question-context pairs, $(q^a_i, \textbf{c}^a_i)$ and $(q^b_i, \textbf{c}^b_i)$, generated from the same source document $d_i$, better aligns with the hierarchical biomedical knowledge embodied in Medical Subject Headings~\cite{mesh} (MeSH). 
The agent employs a two-stage approach: an initial rule-based evaluation leveraging the MeSH ontology to create high-quality preference labels, followed by learning an automatic evaluator via LLM fine-tuning.

\subsubsection{Knowledge-Driven Fine-Tuning}
To enable evaluation without relying on human-annotated preference data, we propose a knowledge-guided, rule-based cold-start method to compare candidate question-context pairs. 
This approach leverages the structural and semantic information from the MeSH ontology to generate high-confidence preference labels at scale.
Given a source document $d_i$ and two candidate pairs $(q^a_i, \textbf{c}^a_i)$ and $(q^b_i, \textbf{c}^b_i)$, we can obtain MeSH terms\footnote{Our dataset comprises research articles from PubMed, each accompanied by its associated MeSH terms for both the document and its context.} from $d_i$, $\textbf{c}^a_i$, and $\textbf{c}^b_i$. 
Referring to Lin's approach~\cite{sim_lin}, for given context set $\textbf{c}_i$, the knowledge hierarchy similarity to $d_i$ is computed as: 
\begin{equation}
\Bar{S}^j_i = \frac{1}{|d_i|\, |\textbf{c}_i|} \sum_{m^x \in d_i} \sum_{m^y \in \textbf{c}_i} \frac{2 \times IC(\Lambda(m^x, m^y))}{IC(m^x) + IC(m^y)},
\end{equation}
where $IC(m)$ is the information content for MeSH term $m$, and $\Lambda(m^x, m^y)$ is their lowest common ancestor.
For any structured subject term $m$, we first calculate its information content as:
\begin{equation}
IC(m) = -\log(\frac{freq(\mathcal{M}(m))}{n_{terms}}),
\end{equation}
where $\mathcal{M}(m)$ denotes the set of all descendants of MeSH term $m$, and $n_{terms}$ represents the total number of MeSH terms in the corpus. 
This score reflects how well the candidate context’s MeSH terms align with those of the source document. A higher $\bar{S}^j_i$ indicates that the candidate context is more semantically consistent with the document's knowledge hierarchy.
Once the similarity scores for each candidate pair have been calculated, we select the one with the higher score:
\begin{equation}
y_i = \arg\max_{j \in \{a,b\}} \Bar{S}^j_i
\end{equation}
In other words, the candidate pair whose context set exhibits a stronger semantic match to the source document is preferred.
These cold-start preference labels, obtained in a rule-based and knowledge-guided manner, provide a large-scale and high-quality supervision signal. This signal can then be used to bootstrap a subsequent LLM-based evaluation agent.

\subsubsection{Evaluation Agent Preference Learning}
With the knowledge-guided cold-start preference labels as supervision, we further train a large language model $\phi$ to serve as an automatic evaluation agent. 
Given a training set of triplets $\{(d_i, (q^a_i, \textbf{c}^a_i), (q^b_i, \textbf{c}^b_i), y_i)\}$, where $y_i$ indicates which candidate pair is preferred according to the MeSH-based scoring, the LLM is optimized to predict the correct preference.
Formally, let $P_{\phi}(y_i | d_i, q^a_i, \textbf{c}^a_i, q^b_i, \textbf{c}^b_i)$ denote the predicted probability that the $y_i$-th pair is preferred. The LLM is trained by minimizing the negative log-likelihood:
\begin{equation}
\mathcal{L}_{\mathrm{eval}}(\phi) = - \frac{1}{N} \sum_{i=1}^N \log P_{\phi}(y_i | d_i, q^a_i, \textbf{c}^a_i, q^b_i, \textbf{c}^b_i),
\end{equation}
where this process enables the Question Evaluation Agent to learn preference judgment criteria derived from domain knowledge, without requiring human-annotated data.

\subsubsection{Automatic Evaluation and Deployment}
Once fine-tuned, the LLM-based evaluation agent can be deployed to automatically assess the relative quality of any two candidate question-context pairs generated from the same document. Given $(d_i, (q^a_i, c^a_i), (q^b_i, c^b_i))$, the agent predicts which pair is better aligned with biomedical knowledge, thus supporting scalable and efficient data distillation for downstream tasks.
This learned evaluation mechanism inherits the domain expertise embedded in the MeSH-based rules and generalizes to more complex or subtle cases, enabling robust multi-agent collaboration in the m-KAILIN framework.

\subsection{Answer Generation Agent}
Once the optimal question-context pair is identified, the Answer Generation Agent $\varphi$ generates high-quality answers using advanced LLMs such as GPT-4o~\cite{achiam2023gpt}. 
For each input $(q_j, \textbf{c}_j)$, this agent produces an answer $\varphi(q_j, \textbf{c}_j)\rightarrow a_j$, facilitating the construction of comprehensive QA datasets for downstream training. 

\subsection{Agentic Collaborative Framework}
This stage integrates outputs from the multiple agents to construct datasets, and further optimizes the question generator through direct preference optimization~\cite{DPO} (DPO):

\begin{figure}[!h]
    \centering
    \includegraphics[width=0.45\linewidth]{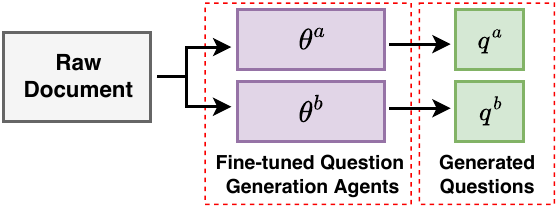}
    \caption{The two distinct Question Generation Agents will generate different question by given the raw documents.}
    \label{fig:method1}
\end{figure}

\smallskip
\subsubsection{Initialize Two Distinct Question Generation Agents} As depicted in Figure~\ref{fig:method1}, m-KAILIN framework will first initialize two Question Generation Agents. 
Specifically, one of them, $\theta^a$, will be based on a domain-specific LLM, e.g., BioMistral~\cite{labrak2024biomistral}.
And the other $\theta^b$ will be based on a more general yet powerful LLM, e.g., Llama-3~\cite{llama3_1}. 
Each of them will follow the same fine-tuning step introduced in Section~\ref{finetune-os} and be able to generate quality questions. 
We want to highlight that the reason we adopt two distinct Question Generation Agents is to enhance diversity and robustness in question generation.
The domain-specific agent (e.g., BioMistral) is expected to capture fine-grained biomedical knowledge and terminology, ensuring that generated questions are highly relevant to the biomedical context.
Meanwhile, the general-purpose agent (e.g., Llama-3) possesses broader reasoning and linguistic capabilities and may generate questions from more diverse perspectives.
By leveraging both agents, m-KAILIN can comprehensively explore the question space over biomedical documents, increasing the likelihood of producing high-quality, informative, and non-redundant question candidates.

\begin{figure}[!h]
    \centering
    \includegraphics[width=0.5\linewidth]{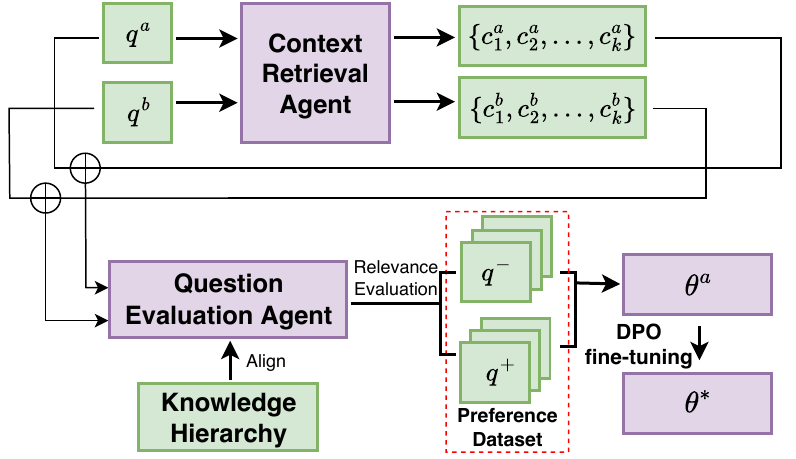}
    \caption{The generated question pairs will be evaluated based on how well their retrieved contexts align with the knowledge hierarchy of the raw document.}
    % \vspace{-0.3cm}
    \label{fig:method2}
\end{figure}

\smallskip
\subsubsection{Retrieve Context of Generated Question} 
The outputs from both agents will be used in the subsequent context retrieval and evaluation steps, allowing the downstream modules to select the most suitable question-context pairs for the construction of high-quality training datasets.
Specifically, each pair of generated questions will feed into the Context Retrieval Agent to obtain their top-$k$ relevant documents as supplementary context (as depicted in the top region of Figure~\ref{fig:method2}). 

\smallskip
\subsubsection{Preference Dataset Construction} 
By utilizing the Evaluation Agent for each generated pair from two Question Generation Agents, we select the question-context pair with the better alignment as the positive sample ($q^+_i$), and the other as the negative ($q^-_i$), forming a preference dataset $\mathcal{P}$ (highlighted by the red dashed line in Figure~\ref{fig:method2}).
This preference dataset forms the supervision signal for optimizing the question generation agent in the next step. 

\smallskip
\subsubsection{Direct Preference Optimization} 
The Direct Preference Optimization (DPO) objective explicitly aligns the model’s output distribution with automated preference signals.
Given a set of preference data $\{(d_i, q^+_i, q^-_i)\}$, where $q^+_i$ is the preferred question over $q^-_i$ for document $d_i$, DPO fine-tunes the general LLM $\theta^b$ to maximize the likelihood of generating preferred questions while minimizing the likelihood of dispreferred ones.
Formally, the DPO objective is:
\begin{align}
\mathcal{L}_{\mathrm{DPO}}(\theta) = 
    - \frac{1}{N} \sum_{i=1}^N 
    \log 
    \frac{
        \exp\left(\beta \log P_\theta(q_i^+|d_i)\right)
    }
    {\begin{aligned}
        &\exp\left(\beta \log P_\theta(q_i^+|d_i)\right) \\
        &+ \exp\left(\beta \log P_\theta(q_i^-|d_i)\right)
    \end{aligned}}
\end{align}
where $\beta$ is a temperature scaling factor. 
The optimized model $\theta^*$ is thus directly steered towards generating questions that align with preference signals and can serve as the basis for constructing high-quality synthetic datasets in the final step.

\begin{figure}[!h]
    \centering
    \includegraphics[width=0.5\linewidth]{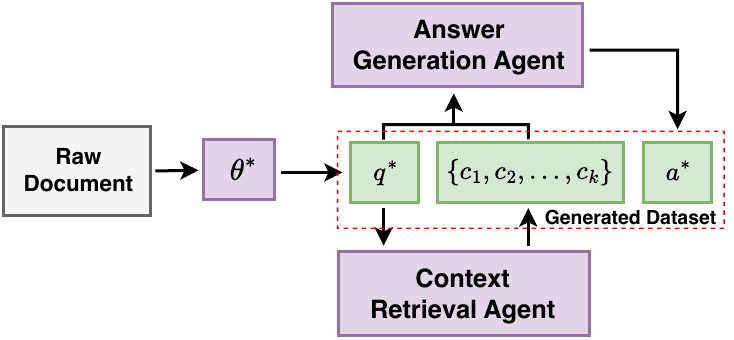}
    \caption{The fine-tuned Question Generation Agent will collaorated with Answer Generation Agent and Context Retrieval Agent to build the training corpora dataset.}
    \vspace{-0.3cm}
    \label{fig:method3}
\end{figure}

\smallskip
\subsubsection{Training Corpus Dataset Construction} 
As shown in Figure~\ref{fig:method3}, the optimized Question Generation Agent $\theta^*$ produces new question-context pairs from raw documents. 
After that, the Answer Generation Agent $\varphi$ provides answers, resulting in two types of ideal datasets: $\mathcal{I}_{\textsc{cpt}}= \{(q_j^*, \textbf{c}_{j})\}_{j=1}^N$, for continued pre-training (CPT), consisting of question-context pairs.
$\mathcal{I}_{\textsc{sft}} = \{(q^*_j, \textbf{c}_{j}, a^*_j)\}_{j=1}^N$, for supervised fine-tuning (SFT), consisting of question-context-answer triples.
This collaborative multi-agent pipeline enables scalable, high-quality question answering data construction without reliance on manual annotation.

\subsection{Training for Downstream Tasks}

Finally, the constructed high-quality training corpus is employed to train a target large language model for biomedical question answering in two sequential stages. For instance, we denote the downstream target model as BioMistral~\cite{labrak2024biomistral}.

\begin{figure}[!h]
    \centering
  \finding[\textbf{Prompt 1}: The prompt for continuous pre-training.]{ \medskip To address the challenges for the biomedical field, I reviewed the paper titled: \colorbox{yellow!20}{\{Title\}}, \colorbox{yellow!20}{\{Context\}}.\\[1ex]
    Motivated by this study, I conducted a literature review to gather additional resources and contextualize its findings. During this process, I identified the following key materials: \colorbox{yellow!20}{\{Retrived Context\}}.\\[1ex]
    Reflecting on these insights, I formulated the following research question: \colorbox{green!20}{\{Question\}}.
  }
\caption{The prompt for continuous pre-training.}
% \vspace{-0.6cm}
\label{fig:Prompt_CPT}
\end{figure}

\begin{figure}[!h]
  \finding[\textbf{Prompt 2:} The prompt for PubMedQA inference.] {\medskip Please analyze the information in the title and context in the field of biomedical and generate a question:\\
    \textbf{Title:}
    \colorbox{yellow!20}{\{Title\}}\\
    \textbf{Context:}
    \colorbox{yellow!20}{\{Context\}}\\
    \textbf{Response:}
    \colorbox{green!20}{Question}
  }
\caption{The prompt for PubMedQA inference.}
% \vspace{-0.2cm}
\label{fig:Prompt_PMQA}
\end{figure}

\smallskip
\subsubsection{Continuous Pre-training} The target LLM is first further pre-trained on the heuristic question-context pairs $\mathcal{I}_{\textsc{cpt}} = \{ (q_j, \textbf{c}_j) \}_{j=1}^N$ generated by the optimized question generator. 
This stage aims to improve the model’s foundational biomedical knowledge and enhance its ability to understand biomedical contexts and question styles.
The prompt of this stage is given as Figure~\ref{fig:Prompt_CPT}.

% \begin{figure*}[h!]
%     \centering
% \includegraphics[width=0.65\textwidth]{Fig/prompt_CPT_with_io.pdf} 
%     \caption{The prompt for continuous pre-training.}
%     \label{fig:Prompt_CPT}
% \end{figure*}

\smallskip
\subsubsection{Supervised Fine-tuning} Subsequently, the model is fine-tuned on question-context-answer triples $\mathcal{I}_{\textsc{sft}} = \{ (q_j, \textbf{c}_j, a_j) \}_{j=1}^N$, as well as optionally on benchmark QA datasets such as PubMedQA PQA-A~\cite{jin2019pubmedqa} training set.
As shown in Figure~\ref{fig:Prompt_PMQA}, this prompt is designed to match the downstream QA format. This stage explicitly optimizes the model for accurate answer generation and robust question understanding.

% \begin{figure*}[h!]
%     \centering
% \includegraphics[width=0.65\textwidth]{Fig/prompt_PMQA_with_io.pdf} 
%     \caption{The prompt for PubMedQA inference.}
%     \label{fig:Prompt_PMQA}
% \end{figure*}

\smallskip
\subsubsection{Summary of the Training Procedure} 
The two-stage training pipeline ensures that the trained LLM can effectively leverage both the large-scale synthetic data from m-KAILIN and high-quality human-annotated benchmarks, substantially boosting downstream biomedical QA performance. 
In practice, the continuous pre-training step adopts the standard language modeling objective, while the supervised fine-tuning step minimizes the cross-entropy loss between the model’s predicted answers and ground-truth answers.

%% file: 4.experiment.tex
\subsection{Experimental Setups} 
\subsubsection{Base Models} 
We utilized the llama-series LLM as base models in our primary experiment, while also incorporating BioMistral~\cite{labrak2024biomistral} as the backend of the Question Generation Agent for building the preference dataset and fine-tuning the question generator. 
We adopted GPT-4o as the backend of the Answer Generation Agent. 

\smallskip
\subsubsection{Baselines} We conducted a comprehensive evaluation of various open-source models, including LLaMA-2~\cite{touvron2023llama}, LLaMA-3~\cite{llama3_1}, Mistral~\cite{jiang2023mistral}, GLM-4~\cite{glm2024chatglm}, Qwen2.5~\cite{yang2024qwen2}, and Gemma\cite{team2024gemma}, Deepseek~\cite{bi2024deepseek}, as well as proprietary models like GPT-4\cite{achiam2023gpt} (with MedPrompt~\cite{nori2023can,nori2023capabilities}), and Med-PaLM~\cite{Med-Palm-2}. In particular, we focused on models specifically trained for the biomedical domain, such as BioMistral~\cite{labrak2024biomistral}, PMC-LLaMA~\cite{wu2024pmc}, HEAL~\cite{HEAL}, KAILIN~\cite{cai2025knowledge} (with Qwen2.5, GLM-4, and LLama-3 as backend), and MMedLM~\cite{MMed_LM}, to demonstrate the effectiveness. 

\smallskip
\subsubsection{Evaluation Benchmarks and Protocols} 
We conduct a comprehensive evaluation using a suite of benchmark subsets derived from PubMedQA~\cite{jin2019pubmedqa}.
We design a multi-dimensional evaluation protocol that probes complementary aspects of biomedical QA capability, including overall accuracy, temporal generalization, subdisciplinary robustness, reasoning sensitivity, and data scaling behavior.
Specifically, we first evaluate all models on the PubMedQA-Full benchmark, which consists of expert-annotated question–answer pairs grounded in biomedical literature and serves as the primary indicator of end-to-end QA performance. To assess robustness under biomedical knowledge evolution, we further construct chronological benchmarks (PubMedQA-Chrono) by partitioning the PQA-L subset into eight time-based slices according to publication periods, enabling evaluation across distinct biomedical eras.

To examine cross-domain generalization, we derive subdisciplinary benchmarks (PubMedQA-MeSH) by grouping samples according to their dominant Medical Subject Headings~\cite{mesh} (MeSH) terms, resulting in six overlapping subsets that reflect major biomedical research areas. 
In addition, following prior work, we distinguish between reasoning-required and question-only inference settings, forming a reasoning-oriented benchmark that isolates the contribution of contextual understanding and biomedical reasoning.

Together, these benchmark settings provide a controlled yet comprehensive evaluation suite, allowing us to systematically analyze the effectiveness and robustness of the proposed m-KAILIN framework across multiple dimensions of biomedical question answering.

\smallskip
\subsubsection{Platform Information} 
Each experiment, either utilizing the API query or performed on an Ubuntu 18.04.6 LTS system featuring an AMD EPYC 7742 processor and eight Cambricon MLU-590B, was carried out within a Python 3.11.0 environment using PyTorch 2.1.1.

\begin{figure}[!t]
    \centering
\subfloat[LLMs of fewer than 13B parameters]{
\includegraphics[width=0.55\textwidth]{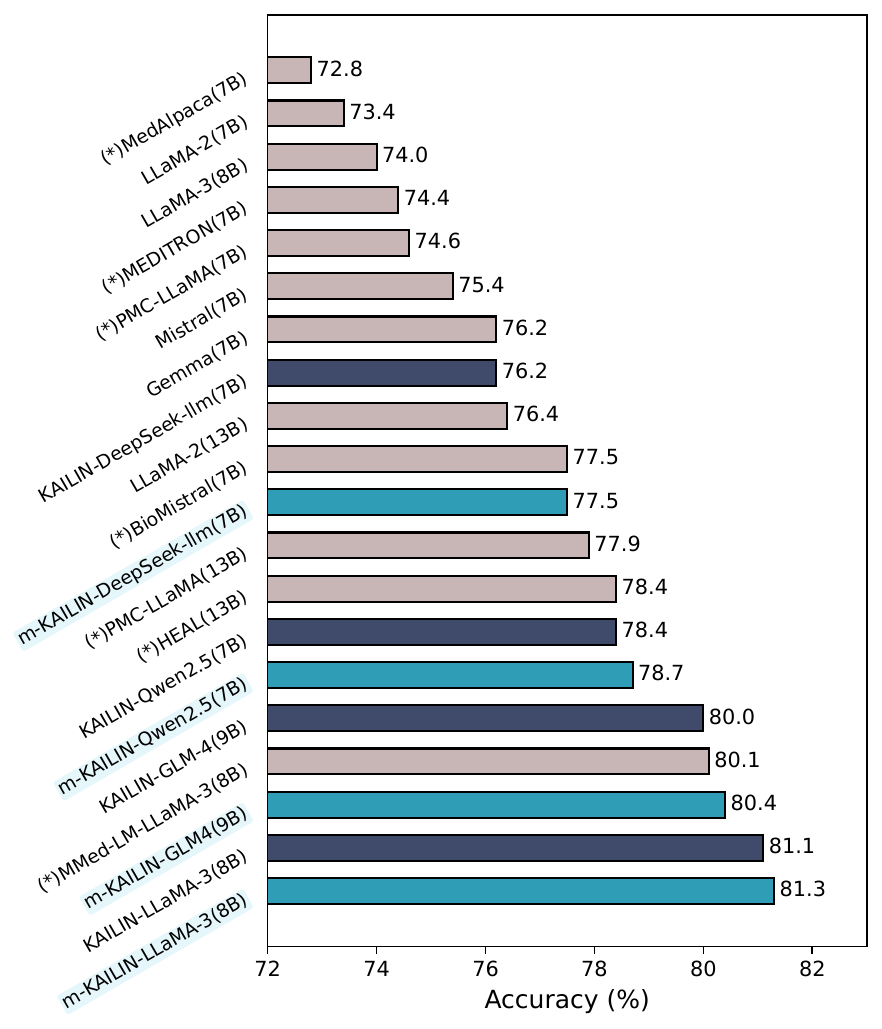}}
        \label{fig:sub1}
\hfill
\subfloat[LLMs of more than 70B parameters]{
\includegraphics[width=0.4\textwidth]{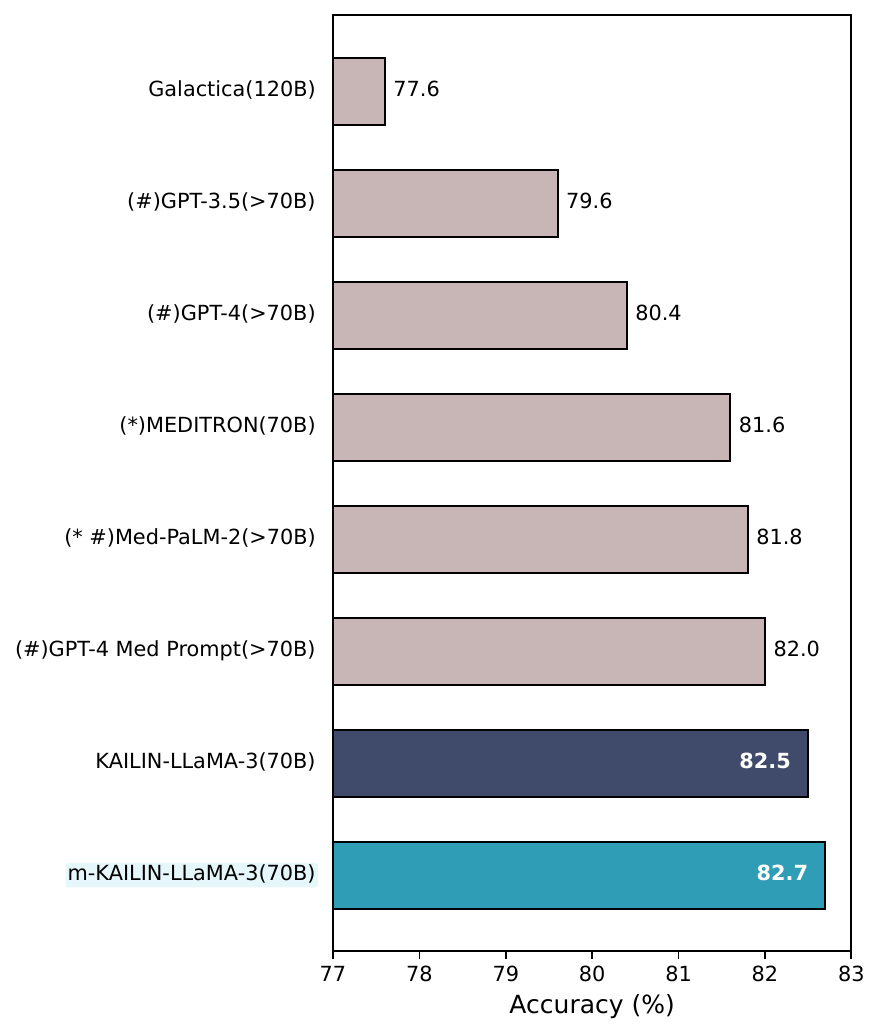}
    \label{fig:sub2}}
\caption{Evaluations (accuracy (\%)) for PubMedQA~\cite{jin2019pubmedqa} problems on our models compared to other open-source models, closed-source models, and domain-specific models. Models marked with * indicate domain-specific biomedical models. The \# symbol denotes closed-source models.}
\label{main_exp}
\end{figure}

\subsection{Main Results and Analysis} \label{exp-main}
As illustrated in Figure~\ref{main_exp}, we benchmarked m-KAILIN and various baseline models across two parameter scales: models with fewer than 13B parameters and those with 70B parameters or greater. 
A clear positive correlation exists between model size and performance: models with over 70B parameters generally surpass smaller models in accuracy. 
However, our results reveal that the application of m-KAILIN narrows this gap, allowing smaller models to achieve competitive performance. 
Notably, m-KAILIN-LLaMA-3 (8B) nearly matches or outperforms several larger domain-specific models, underscoring the potential of architectural and training innovations to offset limitations in model scale partially.
The key driver is the MeSH-based knowledge hierarchy similarity evaluation integrated within Question Evaluation Agent, which substantially enhances a model's ability to comprehend and reason about the complex terminologies and conceptual structures of biomedical texts. 
This demonstrates the effectiveness of our method in adapting smaller, general-purpose models to specialized domains—a critical advantage given the rapid iteration cycles in large model development.
In the large-model group, m-KAILIN significantly boosts the performance of LLaMA-3-70B, enabling it to outperform powerful closed-source systems such as GPT-4 with MedPrompt and Med-PaLM-2, despite their larger parameter counts. This suggests that, while increasing model size typically yields better results, targeted preference alignment using knowledge hierarchy evaluation can enable smaller or open-source models to close the gap or even exceed the performance of much larger models, thus offering substantial efficiency and accessibility benefits.
In summary, the m-KAILIN framework enables both small and large general-purpose models to achieve state-of-the-art performance on domain-specific biomedical tasks.
These experimental results also validate the feasibility of leveraging the m-KAILIN framework to automatically extract high-quality training data from raw textual sources.

\subsection{Ablation Studies of each Technical Component} \label{exp-ablation}
The main motivation of this ablation study is to systematically dissect the contributions of each key component in our proposed multi-agent framework for biomedical corpus construction. 
Specifically, we aim to understand how multi-agent diversity (in question generation), domain-specific retrieval, and knowledge-aware evaluation impact the quality of the generated datasets and the downstream performance of LLMs.

\begin{figure}[!h]
    \centering
\includegraphics[width=0.45\linewidth]{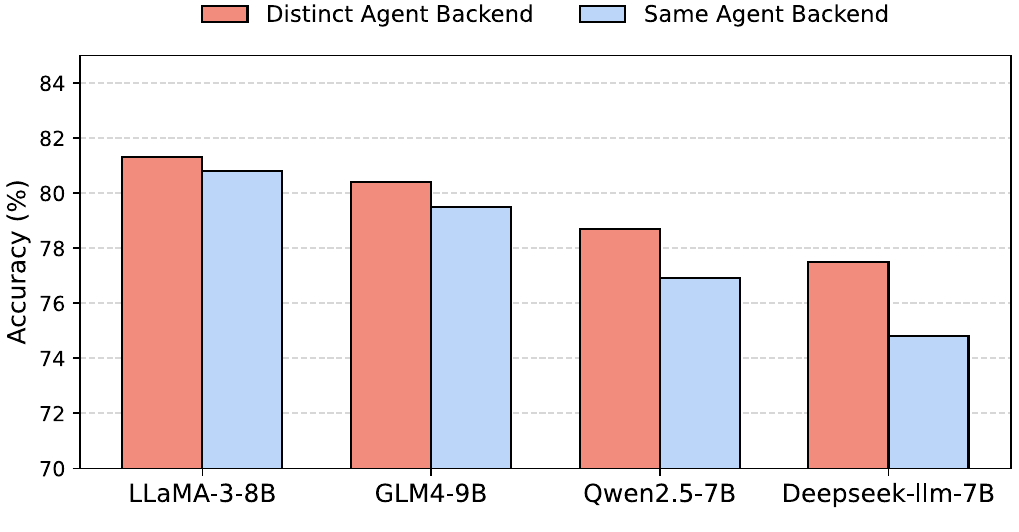}
    \caption{Ablation study on the question generation agent backend. Comparison between using two identical general-purpose agents (Same Agent Backend) and combining a general-purpose agent with a domain-specific agent (Distinct Agent Backend).}
    \label{fig:abl-dis}
\end{figure}

\subsubsection{Ablation Study on Question Generation Agent}
We examine whether leveraging heterogeneous agents for question generation—namely, combining one general-purpose and one domain-specific agent (Distinct Agent Backend)—offers advantages over using two identical, general-purpose question generation agents (Same Agent Backend). The rationale is that agent diversity may enhance the richness and coverage of generated questions by introducing complementary knowledge and styles.

As shown in Figure~\ref{fig:abl-dis}, across all evaluated LLMs, the Distinct Agent Backend consistently outperforms the Same Agent Backend. For example, LLaMA-3-8B achieves 81.3\% accuracy with the distinct setup, compared to 80.8\% with identical agents; similar trends are observed for GLM4-9B, Qwen2.5-7B, and Deepseek-llm-7B. This demonstrates that combining a general and a domain-specific QG agent leads to more diverse and higher-quality question-answer pairs, thereby enhancing downstream QA performance. The results suggest that agent heterogeneity is a key driver for generating a richer and more challenging biomedical corpus.

\begin{figure}[!h]
    \centering
\includegraphics[width=0.45\linewidth]{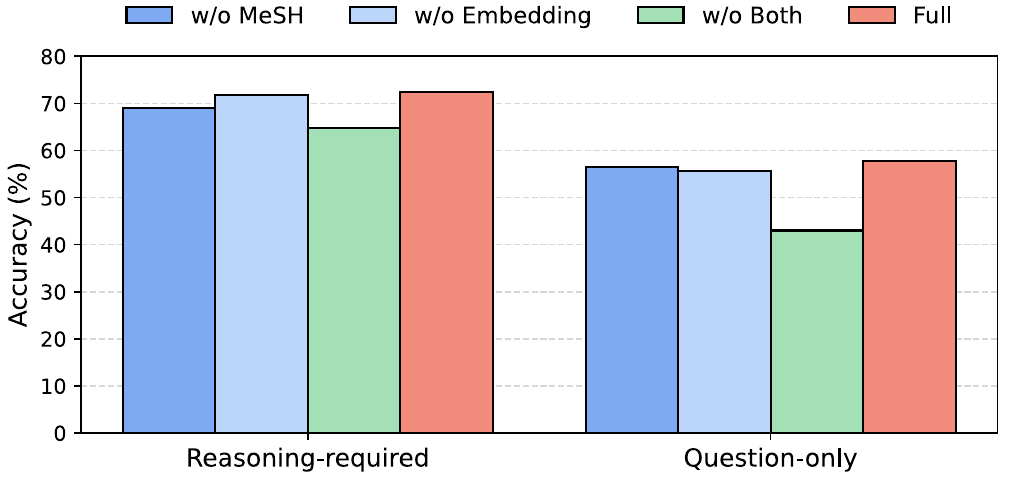}
    \caption{Evaluations of the overall experimental results of the ablation study. The reasoning-required and question-only are both inference settings in PubMedQA study.}
    \label{table_exp_ablation}
\end{figure}

\subsubsection{Ablation Study on  Context Retrieval Agent and Question Evaluation Agent}

To further investigate the roles of domain-specific knowledge and retrieval in our framework, we conduct ablation experiments focusing on the evaluation and retrieval agents. Specifically, we consider two ablation scenarios:

\begin{itemize}
    \item w/o \textbf{MeSH}: The evaluation agent does not utilize the MeSH-based knowledge hierarchy; instead, it directly employs a general-purpose LLM for evaluation without hierarchical biomedical guidance.
    \item w/o \textbf{Embedding}: The retrieval agent does not use a domain-adapted embedding model; instead, it relies on a general model such as BERT without further fine-tuning on biomedical data.
\end{itemize}

Figure~\ref{table_exp_ablation} shows the performance across different configurations. Removing MeSH (w/o MeSH) leads to a substantial decrease in accuracy, especially in the reasoning-required setting. 
This suggests that biomedical hierarchical knowledge—provided by MeSH—is crucial for the evaluation agent to accurately judge the relevance and quality of context-question pairs. Without this structured guidance, the general LLM struggles to evaluate biomedical content effectively, resulting in lower overall performance.
In the w/o Embedding setting, where the retrieval agent uses general-purpose BERT embeddings instead of a fine-tuned, domain-specific model, performance also drops.
This indicates that domain-adapted embeddings significantly enhance the retrieval of relevant biomedical contexts. General embeddings may miss domain-specific nuances, leading to suboptimal context selection and thereby affecting downstream QA accuracy.
Notably, when both MeSH and domain-adapted embedding are removed (w/o Both), there is a pronounced performance degradation (64.8\% in reasoning-required, 43\% in question-only), confirming that both hierarchical knowledge in evaluation and domain adaptation in retrieval play complementary, essential roles.
Overall, these results demonstrate that integrating domain-specific components—both in evaluation and retrieval—is vital for constructing a high-quality biomedical corpus and maximizing LLM downstream performance.

\subsection{Hyperparameter Experiment}
In the m-KAILIN framework, the number of retrieved documents in context, Top-$k$, is a highly influential parameter. 
This parameter directly impacts the balance between the comprehensiveness, richness of context information, and the upper limit of contextual understanding of models. 
We conducted hyperparameter experiments to study the effect of retrieving different numbers of $k$ documents as context on model performance. 
As shown in Figure \ref{hyperparam_exp}, the model exhibited the best performance when $k$ was set to 4. 
This optimal $k$ allows the model to gain a more comprehensive understanding of the overall information relevant to the question in the biomedical domain, improve alignment, and reduce noise. 
We use this optimal setting for each of the experiments.

\begin{figure}[!h]
\centering
\begin{tikzpicture}
  \begin{axis}[
    width=8.5cm,              % 宽度
    height=4cm,             % 高度，正方形
    grid=major,
    xlabel={$k$ (Retrieval Number)},
    ylabel={Accuracy (\%)},
    ymin=68, ymax=74,
    xmin=1, xmax=9,
    xtick={2,4,6,8},
    ytick={68,69,70,71,72,73,74},
    legend style={at={(0.8,0.3)},anchor=north, font=\small},
    every axis plot/.append style={thick,mark=*}
    ]
    \addplot[
      color=blue
    ] coordinates {
      (2, 69.4)
      (4, 72.8)
      (6, 72.7)
      (8, 72.4)
    };
    \addlegendentry{Accuracy}
  \end{axis}
\end{tikzpicture}
\caption {Evaluation(accuracy(\%)) on PubMedQA using continuous pre-training datasets of retrieving different Top-$k$ documents.}
% \vspace{-0.6cm}
\label{hyperparam_exp}
\end{figure}

\begin{figure*}[!h]
  \includegraphics[width=\linewidth]{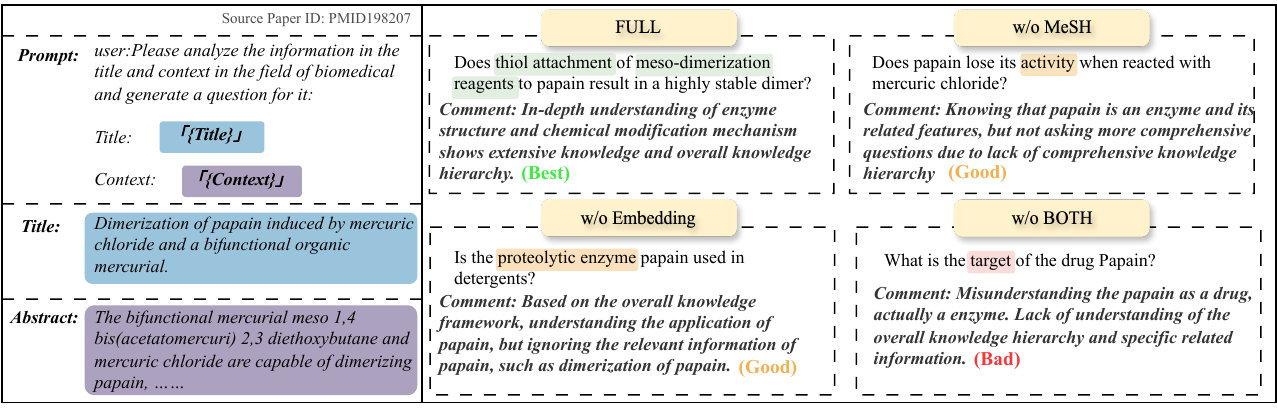}
  \caption {A case study comparing the generated questions under different experimental settings in the ablation study.}
  % \vspace{-0.6cm}
\label{case_study}
\end{figure*}

\subsection{Case Study of Generated Corpus}
From the specific cases illustrated in Figure~\ref{case_study}, it is evident that the model can recognize "activity" in relation to papain's function as an enzyme, demonstrating a general understanding facilitated by domain cues. However, without comprehensive and structured biomedical knowledge, the model still exhibits limitations in generating integrative questions that require a deeper or more holistic perspective.
Performance differences under the question-only inference setting further highlight these shortcomings. In particular, when the embedding model is not fine-tuned on biomedical data (i.e., using a general-purpose embedding), the model struggles to capture question-specific nuances, resulting in significantly reduced performance when relevant context is unavailable.
Our case analysis also reveals the pivotal role of the MeSH-based knowledge hierarchy. When MeSH guidance is present, the model is able to correctly interpret papain as a "proteolytic enzyme" within the broader biomedical context. However, the absence of fine-grained, domain-specific knowledge—especially when leveraging only general LLMs or embeddings—leads to missed opportunities in question generation, such as failing to inquire about detailed aspects like enzyme activity or dimerization.
These findings underscore the necessity of integrating both structured biomedical ontologies (e.g., MeSH) and domain-adapted retrieval mechanisms in our multi-agent framework. Only through this combination can the model attain both the breadth and depth required for high-quality, knowledge-rich biomedical question generation and evaluation.

\subsection{Scaling Law of Dataset Distillation} 

We conducted experiments using datasets of varying scales for continuous pre-training, followed by supervised fine-tuning under consistent experimental settings, to evaluate the impact of different data volumes on model performance. 
In this study, we selected the Llama-3-8B and Llama-3-70B models as base models, to verify the impact of varying dataset scales on models of different sizes. 

\begin{figure}[!h]
\centering
  % \includegraphics[width=0.85\linewidth]{Fig/incremental_exp_new_2.pdf}
  % \caption {Evaluation(accuracy(\%)) of Llama-3-8B and Llama-3-70B on PubMedQA using datasets of varying scales, from 20,000 (20K) to 100,000 (100K).}
  % % \vspace{-0.6cm}
  \begin{tikzpicture}
  \begin{axis}[
    width=10.5cm,              % 宽度
    height=4cm, 
    grid=major,
    xlabel={Scale ($\times10^3$)},
    ylabel={Accuracy (\%)},
    xtick={20,30,40,50,60,70,80,90,100},
    ymin=74, ymax=84,
    ytick={72,74,76,78,80,82,84},
    legend style={at={(0.5,1.05)},anchor=south, font=\small},
    tick label style={font=\small},
    every axis plot/.append style={thick, mark=*},
    cycle list name=color list,
  ]
    % m-KAILIN(LLaMA-3-8B)
    \addplot[color=blue,mark=*] coordinates {
      (20,78.1)
      (80,81.2)
      (100,82.5)
    };
    \addlegendentry{m-KAILIN-LLaMA-3 (8B)}

    % m-KAILIN(LLaMA-3-70B)
    \addplot[color=red,mark=square*] coordinates {
      (20,74.4)
      (40,75.5)
      (60,78.3)
      (80,79.9)
      (100,81.1)
    };
    \addlegendentry{m-KAILIN-LLaMA-3  (70B)}
  \end{axis}
\end{tikzpicture}
\caption {Evaluation (accuracy(\%)) of Llama-3-8B and Llama-3-70B on PubMedQA using datasets of varying scales, from 20,000 (20$\times10^3$) to 100,000 (100$\times10^3$).}
\label{incremental_dataset_exp}
\end{figure}

% We established experimental groups with dataset scales of 20K, 40K, 60K, 80K, and 100K samples for the former, and 20K, 80K, and 100K samples for the latter.
As shown in Figure \ref{incremental_dataset_exp}, we observed that the larger LLaMA-3-70B model demonstrates overall better performance compared to the smaller LLaMA-3-8B model. 
We attributed this advantage to its larger parameter scale, which provides superior feature capture and generalization capabilities.
% As shown in Figure \ref{incremental_dataset_exp}, the larger LLaMA-3-70B model, due to its superior feature-capturing capability and generalization ability, demonstrated an overall better performance than the smaller LLaMA-3-8B. 
% On datasets of varying sizes, both models showed a trend of performance improvement as the dataset size increased. 
% The larger datasets distilled by KAILIN are more capable of approximating the overall distribution of the original dataset. 
We also observed that on datasets of varying sizes, both models showed a trend of performance improvement as the dataset size increased. 
The underlying driver is that larger datasets distilled by m-KAILIN are better able to represent a broader range of biomedical knowledge due to a greater number of documents being comprehensively analyzed.
% approximate the overall distribution of the raw dataset. 
This phenomenon indicates that the m-KAILIN framework indeed produces higher-quality datasets when distilling larger-scale datasets. 
Moreover, we found that while smaller models may not outperform larger models, they seem to benefit more from progressively larger distilled datasets. 
We attribute this phenomenon to the degree of alignment between task complexity and model scale. 
This phenomenon indicates that larger models, due to potential knowledge redundancy, may gain less from incremental datasets on a specific task.

% \begin{figure*}[!t]
%   \includegraphics[width=\linewidth]{Fig/hyperparam_exp.pdf}
%   \caption {Evaluation(accuracy(\%)) on PubMedQA using continuous pre-training datasets of retrieving different Top-$k$ documents.}
% \label{hyperparam_exp}
% % \vspace{-0.6cm}
% \end{figure*}

\begin{figure*}[!t]
  % \centering
  \begin{minipage}{0.7\textwidth} 
    % \centering
    \includegraphics[width=\textwidth]{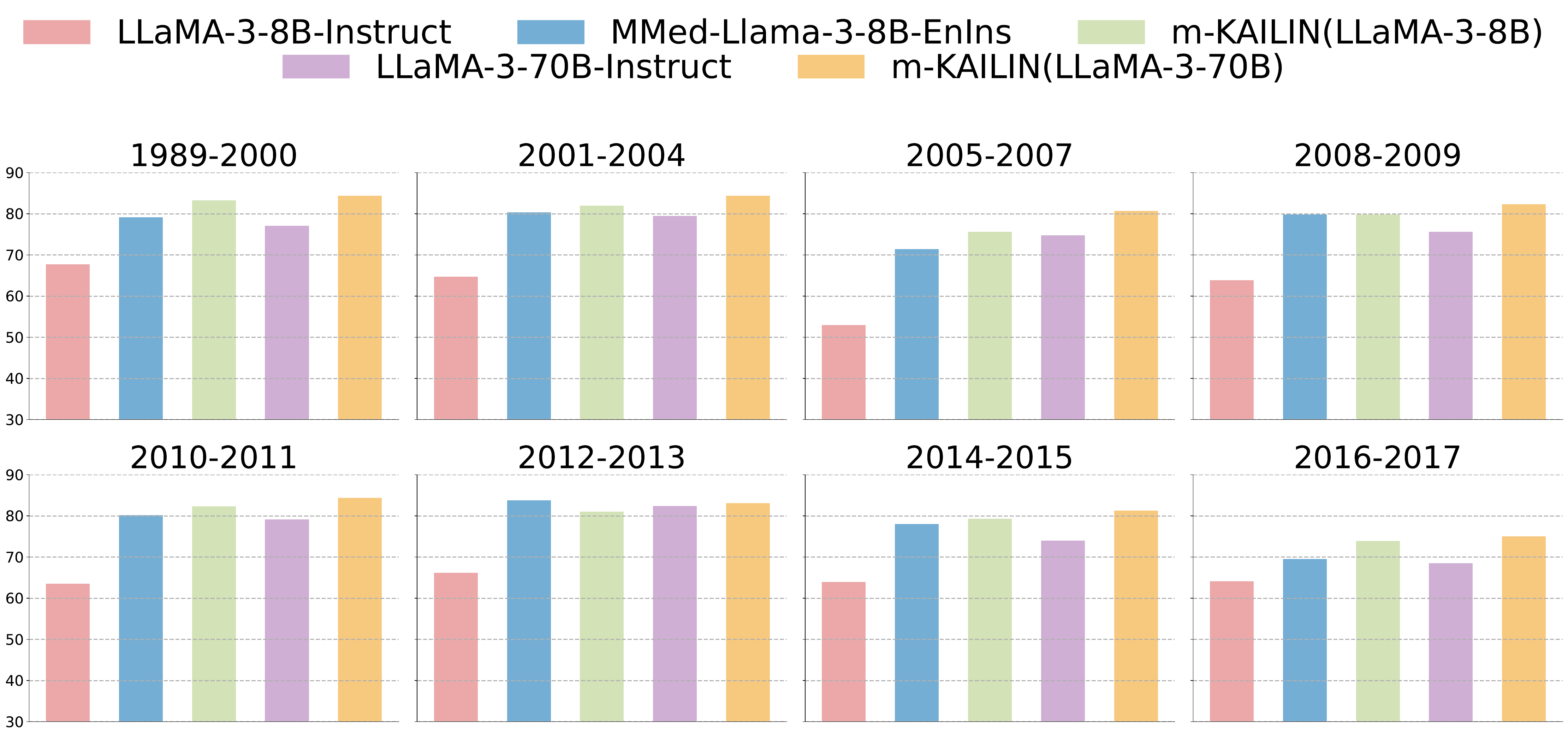} 
  \end{minipage}%
  \begin{minipage}{0.3\textwidth} 
    % \centering
    \scriptsize 
    \renewcommand{\arraystretch}{1.7} 
    \begin{tikzpicture}
    \pie[text=legend,
         radius=1.6,
         sum=auto,
         after number = \%,
         color={cb1, cb2, cb3, cb4, cb5, cb6, cb7, cb8}]
      {9.6/1989-2000,
       12.2/2001-2004,
       11.9/2005-2007,
       11.9/2008-2009,
       9.6/2010-2011,
       14.8/2012-2013,
       15.0/2014-2015,
       9.2/2016-2017}
  \end{tikzpicture}
  % \caption{Proportion of Time Spans}
  \end{minipage}
  \caption{Evaluations (accuracy (\%)) for different chronological subsets of problems in PubMedQA PQA-L on our models compared to other representative models.}
  \vspace{-0.6cm}
  \label{time_span_exp}
\end{figure*}

\begin{figure*}[!t]
  % \centering
  \begin{minipage}{0.70\textwidth} 
  % \raggedright
    \includegraphics[width=\textwidth]{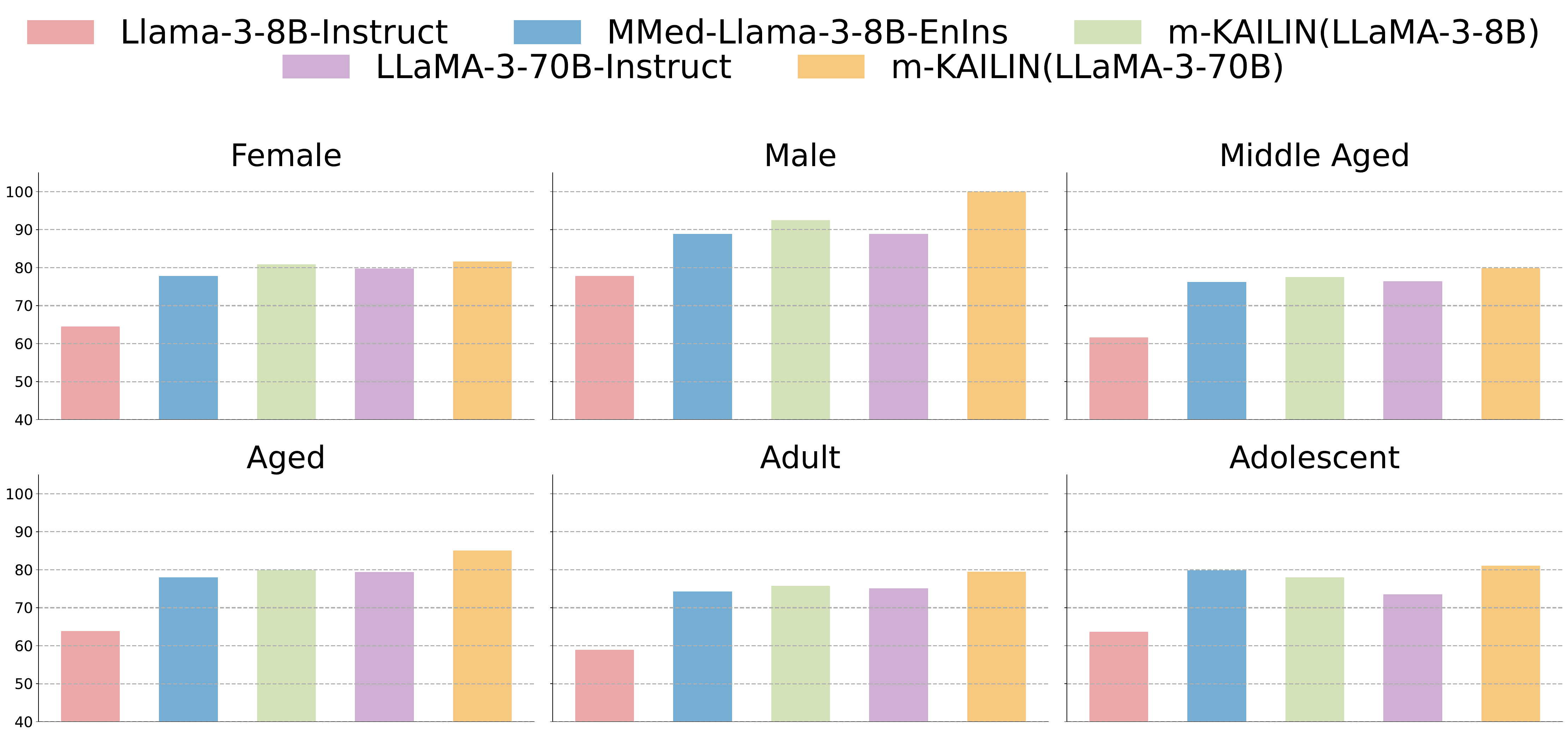} 
  \end{minipage}%
  \begin{minipage}{0.23\textwidth} 
    % \centering
    \scriptsize 
    \begin{tikzpicture}[xshift=-3.25cm]
      \pie[
        text=legend,
        radius=1.45,
        sum=auto,
        after number=,
        color={cb1, cb2}
      ]{
        785/Female,
        703/Male
      };
\end{tikzpicture}
  % Age Distribution (bottom)
  \begin{tikzpicture}[yshift=-3.25cm]
    \pie[
      text=legend,
      radius=1.45,
      sum=auto,
      after number=,
      color={cb3, cb4, cb5, cb6}
    ]{
      542/Middle Aged,
      414/Aged,
      492/Adult,
      204/Adolescent
    };
  \end{tikzpicture}

  \end{minipage}
  \caption{Evaluations (accuracy (\%)) for different MeSH term subsets problems in PubMedQA PQA-L on our models compared to other representative models.}
  \label{topic_exp}
\end{figure*}

\subsection{Robustness toward Different Question Groups}
Our robustness experiments investigate how well the m-KAILIN framework generalizes across different temporal spans and biomedical subdisciplines, reflecting real-world application scenarios that challenge LLMs with both chronological and disciplinary diversity.

\subsubsection{Robustness Under Chronological Question}
As shown in Figure~\ref{time_span_exp}, our agentic framework, especially when combined with MeSH-based evaluation, enables models to maintain consistently high accuracy across documents published at different time periods. 
Compared to strong biomedical baselines such as MMedLM, models distilled with m-KAILIN demonstrate significantly greater robustness to temporal shifts. This improvement stems from two key innovations: (1) the multi-agent setup, which incorporates diverse perspectives during corpus construction, and (2) the MeSH-guided evaluation agent, which provides structured biomedical context, allowing the model to better understand and adapt to evolving domain knowledge over time. 
Notably, larger models exhibit even greater temporal robustness, likely due to their increased capacity to absorb and generalize from temporally diverse data curated by our framework.

\subsubsection{Robustness Under Subdisciplinary Question}
We conduct evaluation on subdisciplinary slices based on major MeSH terms (Figure~\ref{topic_exp}).
The results show that m-KAILIN-trained models outperform their peers across various biomedical topics and demographics. 
This robustness stems from our MeSH-based knowledge-driven design, which leverages MeSH hierarchies during both retrieval and evaluation. 
This design ensures comprehensive coverage and nuanced understanding across biomedical subfields. 
The evaluation agent’s use of MeSH facilitates more relevant and context-aware question-answer generation, systematically exposing the model to the breadth of the biomedical literature's structure. 
% Some variability across demographic terms highlights ongoing challenges. 
Finally, the overall trend affirms the effectiveness of our approach in producing models well-adapted to the complex, hierarchical landscape of biomedical knowledge.

%% file: 5.conclusion.tex
In this work, we present a knowledge-driven, multi-agent framework for scientific corpus distillation tailored to biomedical large language model training. 
By leveraging a collaborative architecture—where specialized agents guided by biomedical ontologies autonomously generate, evaluate, and refine question-answer pairs—our approach addresses the limitations of existing open-source scientific corpora in both scale and quality.
Through extensive experiments, we demonstrate that language models trained on our multi-agent distilled datasets achieve substantial improvements in biomedical question-answering tasks, outperforming both strong open-source and proprietary baselines. 
Our ablation studies further validate the effectiveness and synergy of each agent within the framework. This study highlights the potential of agentic, knowledge-guided corpus construction for advancing biomedical AI, and provides scalable tools and datasets to the community for future research. 